\documentclass[times]{elsarticle}

\usepackage{lineno,hyperref}

\journal{}

\usepackage{multirow}
\usepackage{epstopdf}
\usepackage{url}
\usepackage{color}
\usepackage{array}
\usepackage{amsmath}
\usepackage{subcaption}
\usepackage{algorithm,algorithmic}

\begin{document}

\begin{frontmatter}

\title{Image Fusion via Sparse Regularization with Non-Convex Penalties}

\author[mymainaddress]{N. Anantrasirichai\corref{mycorrespondingauthor}}
\cortext[mycorrespondingauthor]{Corresponding author}
\ead{n.anantrasirichai@bristol.ac.uk}

\author[mymainaddress]{Rencheng Zheng}
\author[mysecondaryaddress]{Ivan Selesnick}
\author[mymainaddress]{Alin Achim}

\address[mymainaddress]{Visual Information Laboratory, University of Bristol, UK}
\address[mysecondaryaddress]{Department of Electrical and Computer Engineering, Tandon School of Engineering, New York University, 11201, USA}

\begin{abstract}
The $L_1$ norm regularized least squares method is often used for finding sparse approximate solutions and is widely used in signal restoration. Basis pursuit denoising (BPD) performs noise reduction in this way. However, the shortcoming of using $L_1$ norm regularization is the underestimation of the true solution. Recently, a class of non-convex penalties have been proposed to improve this situation. This kind of penalty function is non-convex itself, but preserves the convexity property of the whole cost function. This approach has been confirmed to offer good performance in 1-D signal denoising. This paper demonstrates the aforementioned method to 2-D signals (images) and applies it to multisensor image fusion. The problem is posed as an inverse one and a corresponding cost function is judiciously designed to include two data attachment terms. The whole cost function is proved to be convex upon suitably choosing the non-convex penalty, so that the cost function minimization can be tackled by convex optimization approaches, which comprise simple computations. The performance of the proposed method is benchmarked against a number of state-of-the-art image fusion techniques and superior performance is demonstrated both visually and in terms of various assessment measures.
\end{abstract}


\end{frontmatter}


\section{Introduction}
Sparse approximations had a growing influence in signal and image processing for the last two decades~\cite{16,24,25}.
 The common method to find the sparse approximate solution is through the $L_1$ norm regularized least squares method, which is a classical solution to inverse problems. The corresponding cost function $J(x)$ is defined as:
\begin{equation}
J(x)=\frac{1}{2}\left\|y-Ax\right\|_2^2+\lambda\left\|x\right\|_1,\quad \lambda>0.
\label{eq:cost}
\end{equation}

In equation (\ref{eq:cost}), the first term is a quadratic term which is called the data fidelity term, the second term is an $L_1$ norm penalty term. 
The observed signal is $y$ which has been degraded by operator $A$, $x$ is the sparse signal or the sparse representation of the signal to be estimated, $\lambda$ is the regularization parameter, which controls the relative contribution between the data attachment term and the penalty term, and $A$ is a linear degradation function. In particular, 
the basis pursuit denoising method \cite{21,22} performs noise reduction in this way, which is also referred to as the lasso problem \cite{17}, and provides good performance when the signal to be estimated admits a sparse approximation with respect to $A$. Generally, $A$ can be an over-complete dictionary or a transform designed to obtain a sparse representation of signals \cite{2}.

The limitation of $L_1$ norm regularization is the underestimation of the high amplitude components of the signal to be estimated, because of the over-smoothing effect of corresponding proximal operator \cite{Aubert:Mathematical:2002,DurandMila:denoising:2007}. Non-convex regularizers can improve the situation, however, the cost functions are likely to be non-convex and it is difficult to find the global minimizer \cite{15,b:achim2010embc}. Hence, Selesnick proposed a new class of non-convex penalty functions to avoid this problem which have better denoising performance in 1-D signals \cite{3}. The new penalty $\psi_{B}(x)$ is called generalized minimax concave (GMC) penalty which is non-convex itself, but maintains the convexity property of the whole cost function. The cost function is in this case
\begin{equation}
J(x)=\frac{1}{2}\left\|y-Ax\right\|_2^2+\lambda\psi_{B}(x),\quad \lambda>0
\label{eq:cost2}
\end{equation}

\noindent where the GMC penalty is defined in terms of a new multivariate generalization of the Huber function:
\begin{equation}
\psi_B (x)=\left\|x\right\|_1-S_{B}(x).
\end{equation}

The new generalized Huber function is defined as an infimal convolution:

\begin{equation}
S_{B}(x)=\inf_{v\in R^N}\big\{\left\|v\right\|_1+\frac{1}{2}\left\|B(x-v)\right\|_2^2\big\}.
\end{equation}

It is proved in \cite{3} that the whole cost function in (\ref{eq:cost2}) is convex when the matrix $B$ in the generalized Huber function in (4) meets the condition:

\begin{equation}
B^TB\preceq\frac{1}{\lambda}A^TA.
\end{equation}

Here, the above framework is adopted and extended to the case of 2-D signals and subsequently applied in order to solve a multisensor data fusion problem. Multisensor image fusion provides a mechanism to combine multiple images into a single representation that has the potential to aid human visual perception or subsequent image processing tasks. Such algorithms endeavour to create a fused image containing the salient information from each source image without introducing artefacts or inconsistencies.
{A number of applications employ image fusion since they require complementary information in a single image, while the capability of a single sensor employed at a single moment in time is limited by design or observational constraints. Several source modalities are utilised to exploit or emphasize various characteristics, such as type of degradation, texture properties, colours, spectral bands, etc. Applications of image fusion include satellite
imaging, medical imaging, robot vision, monitoring and surveillance.}
 Existing pixel-level fusion schemes {\cite{Li:Pixel:2017}} range from simple averaging of the pixel values of registered images to more complex multiresolution (MR) pyramids , sparse methods~\cite{b:achim08a,b:achim08b}, { and recent deep learning based methods \cite{Liu:deep:2018}}. 

In this paper, we pose the image fusion problem as an inverse one and develop an algorithm based on sparse representations and convex regularization that uses non-convex penalty functions. A second important contribution of this work is that we additionally solve the problem of jointly fusing and deconvolving multisensor images. { That is, not only salient information is brought out from each image input, but at the same time the details are enhanced for better visualisation.}

The remainder of the paper is organized as follows: Section 2 introduces the inverse problem that we address. In Section 3, we demonstrate a theorem that states the necessary condition needed for the matrix $B$ to ensure convexity of the cost function. The solution to the resulting convex optimization problem, based on the forward-backward splitting (FBS) algorithm is described in Section 4. The results of the proposed image fusion algorithm are presented in Section 5, and the conclusions in Section 6. 

\section{Problem Formulation}\label{sec:problem}

The multisensor image fusion problem can be posed by considering the following generative model
\begin{equation}
{y_i} = {\beta_i}{H_i}x + {n_i}, \quad i = 1,...,N
\label{eq:fus_model}
\end{equation}

\noindent {where $y_i$ are the $N$ source images, $x$ is the fused image to be estimated, $\beta_i$ is the sensor gain of sensor $i$, and $H$ is a convolution operator that models the usual degradation process that occurs in any image acquisition system. $n_i$ is noise.}
In our proposed algorithm, image fusion is performed in the wavelet domain and without loss of generality we only consider the problem of fusing two images. Hence, the actual cost function corresponding to (\ref{eq:fus_model}) is:
\begin{equation}
\begin{split} 
J(x)=\frac{1}{2}\left\|y_1-\beta_{1}H_{1}Wx\right\|_2^2+\frac{1}{2}\left\|y_2-\beta_{2}H_{2}Wx\right\|_2^2\\
+\lambda\psi_{B}(x),\quad \lambda>0.
\end{split}
\label{eq:cost_function}
\end{equation}

In equation (\ref{eq:cost_function}), there are two data fidelity terms because there are two source images, $x$ is the wavelet representation of the fused image, $W$ represents the inverse wavelet transform and $\psi_{B}(x)$ is the GMC penalty term mentioned before.  By minimizing the cost function, the sparse approximation of the restored image can be estimated. Through inverse wavelet transform, the final fused image can be generated.

In Selesnick’s paper~\cite{2}, the sparse representation of the 1-D signal is obtained through 1-D discrete Fourier transform. Here for 2-D images, the 2-D multilevel discrete wavelet transform is used to obtain the wavelet domain sparse coefficients of the image as it is commonly known that the wavelet domain is more suitable for image processing than the Fourier domain \cite{b:achim08a,16}. 

The sensor gain is also called the sensor selectivity coefficient. There are many different techniques that can be used to estimate the sensor gain. In this paper, the sensor gain is estimated using the principal component analysis (PCA) method in \cite{4,12}, due to its noise robustness property and calculation efficiency. In this method, the source images are divided into several image patches, and the pixels are regrouped lexicographically, which allows their interpretation as n-variate random variables. The specific sensor gains $\beta_i$  of each pair of patches (assume two source images) are obtained through finding the principal eigenvector of the correlation matrix. We assume that in each image patch, the sensor gain is considered as a constant. After the calculation of the sensor gains for all the image patches, the final sensor gain parameters for each corresponding source image can be obtained.

\section{Regularization With a Nonconvex Penalty}

Having defined the cost function for performing joint image fusion and deconvolution in Eq. (\ref{eq:cost_function}), in this section we turn our attention to the way to set the penalty $\psi_{B}(x)$ in order to maintain convexity of (\ref{eq:cost_function}). Condition (\ref{eq:B}) below imposed upon the matrix $B$ ensures this property.

\noindent\textbf{Theorem:} \textit{If}
\begin{equation}
B^TB=\frac{\gamma}{\lambda}(W^TH_1^T\beta_1^2H_1W+W^TH_2^T\beta_2^2H_2W),\ \ \ 0\le\gamma\le1
\label{eq:B}
\end{equation}

\textit{then the function $J$ defined in (\ref{eq:cost_function}) is convex.}

\noindent The proof of this theorem follows directly from that of \textit{Theorem 1} in ~\cite{3} and is included here for completeness of the presentation.

\noindent\textbf{Proof:} Let $v\in R^N$, and
\begin{align} 
Z_1(x)=\frac{1}{2}\left\|y_1-\beta_{1}H_{1}Wx\right\|_2^2\\ Z_2(x)=\frac{1}{2}\left\|y_2-\beta_{2}H_{2}Wx\right\|_2^2  
\end{align} 

\noindent {Due to the Proposition 12.14 in \cite{14}, we can replace `inf' with `min'.} Then the cost function
\begin{align} 
J(x)&=Z_1(x)+Z_2(x)+\lambda\psi_{B}(x) \notag\\ 
&=Z_1(x)+Z_2(x)+\lambda(\left\|x\right\|_1-S_{B}(x))\notag\\
&=Z_1(x)+Z_2(x)+\lambda\left\|x\right\|_1-\min_{v \in R^N}\big\{\lambda\left\| v\right\|_1+\frac{\lambda}{2}\left\|B(x-v)\right\|_2^2\big\}\notag\\
&=\max_{v \in R^N}(Z_1(x)+Z_2(x)+\lambda\left\|x\right\|_1-\lambda\left\| v\right\|_1+\frac{\lambda}{2}\left\|B(x-v)\right\|_2^2)\notag\\
&=\max_{v \in R^N}(\frac{1}{2}x^TZ_3(x)x+\lambda\left\|x\right\|_1+g(x,v))\notag\\
&=\frac{1}{2}x^TZ_3(x)x+\lambda\left\|x\right\|_1+\max_{v \in R^N}(g(x,v)),
\label{eq:cost_function_prove}
\end{align}

\noindent where $Z_3(x)=W^TH_1^T\beta_1^2H_1W+W^TH_2^T\beta_2^2H_2W-\lambda\ B^TB$, and  { $ g(x,v)= - (y_1 \beta_{1}H_{1} + y_2\beta_{2}H_{2})Wx - \lambda\left\|v\right\|_1-  \lambda x B^T B v + \frac{\lambda}{2}v^T B^T B v)$.}
 
In (\ref{eq:cost_function_prove}), $g(x,v)$ is affine in $x$. The last term $max_{v\in R^N}g(x,v)$ is convex since it is the pointwise maximum of a set of convex functions \cite{14}. Therefore, if $Z_3(x)$ is positive semidefinite, the cost function $J(x)$ will be convex. 
Hence, for the cost function in (\ref{eq:cost_function}) to be convex, the matrix $B$ should satisfy (\ref{eq:B}).

Let us also note that if we only want to solve an image fusion problem, i.e. without deconvolution, the operator $H$ can be simply set to be the identity matrix $(H_1=H_2=I)$ and hence the cost function simply becomes: 
\begin{equation}
J(x)=\frac{1}{2}\left\|y_1-\beta_{1}Wx\right\|_2^2+\frac{1}{2}\left\|y_2-\beta_{2}Wx\right\|_2^2\\
+\lambda\psi_{B}(x).
\label{eq:cost_function2}
\end{equation}

The corresponding matrix $B$ that ensures overall convexity of (\ref{eq:cost_function2}) subsequently becomes:
\begin{equation}
B^TB=\frac{\gamma}{\lambda}(W^T\beta_1^2W+W^T\beta_2^2W)=\frac{\gamma}{\lambda}(W^T(\beta_1^2+\beta_2^2)W).
\end{equation}

The estimated sensor gain is the normalized eigenvalue, hence it meets
\begin{equation}
\beta_1^2+\beta_2^2=1.
\end{equation}

\noindent Then the expression for $B$ can be simplified (when $H_1=H_2=I$) as
\begin{equation}
B=\sqrt{\frac{\gamma}{\lambda}}W,\quad \lambda>0,\quad0\le\gamma\le1.
\label{eq:Bsimple}
\end{equation} 

 \begin{algorithm}
 \caption{Pseudocode of optimization algorithm}
 \label{al:Pseudocode}
 \begin{algorithmic}[1]
 \renewcommand{\algorithmicrequire}{\textbf{Input:}}
 \renewcommand{\algorithmicensure}{\textbf{Output:}}
 \REQUIRE $\rho=\max(1,\gamma/(1-\gamma))$, $\mu:0<\mu<\frac{2}{\rho}$, $K:$ maximum iteration number
 \ENSURE  $x^K$, $v^K$
  \FOR {$i = 0$ to $K$}
  \STATE $w^{(i)}=x^{(i)}-\mu$($W^TH_1^T\beta_1(\beta_1H_1Wx^{\left(i\right)}-y_1)$\\
\ \ \ \ $+W^TH_2^T\beta_2(\beta_2H_2Wx^{(i)}-y_2)$\\
\ \ \ \ $+\gamma(W^TH_1^T\beta_1^2H_1W{(v}^{(i)}-x^{\left(i\right)})$\\
\ \ \ \ $+W^TH_2^T{\beta_2^2H}_2W(v^{(i)}-x^{(i)}))$
\STATE $u^{(i)}=v^{(i)}-\mu\gamma(W^TH_1^T\beta_1^2H_1W(v^{(i)}-x^{(i)})$\\
\ \ \ \ $+W^TH_2^T{\beta_2^2H}_2W(v^{(i)}-x^{(i)}))$ 
 \STATE  $x^{(i+1)}=\text{soft}(w^{(i)},\mu\lambda)$
 \STATE  $v^{(i+1)}=\text{soft}(u^{(i)},\mu\lambda)$
  \ENDFOR
 \end{algorithmic} 
 \end{algorithm}
 
\section{Convex Optimization}
\label{sec:convex}

In the previous section, the proposed cost function was proved to be convex when the matrix $B$ is appropriately chosen as in Eq. (\ref{eq:B}) or (\ref{eq:Bsimple}). In this section, the convex optimization algorithm designed to minimize the corresponding cost function in Eq. (\ref{eq:cost_function}) is introduced. The forward-backward splitting (FBS) algorithm \cite{5,6} is employed. FBS is an iterative algorithm, attempting to minimize functions of the form
\begin{equation}
\min_{x}{f_1\left(x\right)+}f_2\left(x\right)
\end{equation}

where $f_1$ is convex and differentiable with $\rho$-Lipschitz continuous gradient $\nabla f_1$, and $f_2$ can be lower semicontinuous convex. Here we apply the FBS algorithm for the minimization of the cost function introduced above. In fact, the optimization of the cost function $J(x)$ in (\ref{eq:cost_function}) can be rewritten as a saddle-point problem:
\begin{equation}
(x^{opt},v^{opt})=arg\min_{x\in R^N}\max_{v\in R^N}F(x,v),
\end{equation}
\begin{equation}
\begin{split} 
F(x,v)=\frac{1}{2}\left\|y_1-\beta_{1}H_{1}Wx\right\|_2^2+\frac{1}{2}\left\|y_2-\beta_{2}H_{2}Wx\right\|_2^2\\
+\lambda\left\|x\right\|_1-\lambda\left\| v\right\|_1+\frac{\lambda}{2}\left\|B(x-v)\right\|_2^2,\quad \lambda>0.
\end{split}
\end{equation}

Substitute the expression of $B$ in (\ref{eq:B}) into $F\left(x,v\right)$, then $F\left(x,v\right)$ changes into:
\begin{equation}
\begin{split} 
F(x,v)=\frac{1}{2}\left\|y_1-\beta_{1}H_{1}Wx\right\|_2^2+\frac{1}{2}\left\|y_2-\beta_{2}H_{2}Wx\right\|_2^2\\
+\lambda\left\|x\right\|_1-\lambda\left\|v\right\|_1+\frac{\gamma}{2}(W^TH_1^T\beta_1^2H_1W\left\|(x-v)\right\|_2^2 \\
+W^TH_2^T\beta_2^2H_2W\left\|(x-v)\right\|_2^2),\quad \lambda>0.
\end{split}
\label{eq:saddle_point}
\end{equation}

The saddle-point problems belong to the class of monotone inclusion problems and these problems can be addressed by FBS algorithm. The saddle-point $(x^{opt},v^{opt})$ of $F(x,v)$ in equation (\ref{eq:saddle_point}) can be obtained through the following iterative algorithm in Algorithm \ref{al:Pseudocode}, where $\text{soft}(\bullet)$ denotes the soft-threshold function:
\begin{equation}
\text{soft}(x,y)=(\max(|x|-y,0)) \: \text{sign}(x).
\end{equation}

After the iterative algorithm, the sparse approximation of the fused image $x$ can be estimated, and through the inverse discrete wavelet transform, the fused image can be reconstructed.
 
\section{Results and Discussions}
\label{sec:results}

In this section, the proposed image fusion algorithm is tested on four datasets of three types, which are multi-focus, visible/IR, and medical images.
There are multiple established methods in the image fusion literature. {Four} of these are selected for benchmarking within the current study, with the first being the classical wavelet domain image fusion using averaging (Wavelet-WA) \cite{Achim:Complex:2005,23}, the second is a recent image fusion algorithm based on sparse representation and dictionary learning (SRDL) in \cite{11},  the third is a convolutional sparse representation (CSR)-based method \cite{Liu:fusion:2016}, { and the fourth is amethod based on convolutional neural networks (CNN) \citep{Ren:infrared:2018}.}

The wavelet transform employed in this paper uses the Haar wavelet basis.
The choice of regularization parameter, $\lambda$, is an important design consideration. In fact, different image datasets, or various assessment metrics, require different regularization parameters, which in turn requires extensive trial-and-error before setting an optimal value. 
For the other parameters, $\gamma$ was set to $0.8$, $\mu$ was set as $1.9\rho$, where $\rho$ is given by $\rho=\max(1,\gamma/(1-\gamma))$.

\subsection{Image fusion}
In the first set of experiments we were interested to test the performance of the proposed method when performing image fusion alone -- no deconvolution. Six image pairs are employed: multi-focus clocks (512$\times$512 pixels), visible and infrared (IR) images of UN camp (320$\times$240 pixels), Building (812$\times$464), Junction (632$\times$496), Octec (640$\times$ 480), and MRI and CT images of the human head (160$\times$160 pixels). If the image input is colour, it is converted to YCbCr format and only the Y channel is employed. After image fusion, the Cb and Cr channels are combined with the fused image and converted back to RBG format.
We set $\lambda=0.005$, which is relatively small, since in this case the input images are virtually noise free, and GMC regularization does not  effect the result to a large extent. 
For classical multi-focus image fusion,  the degree of defocus blur might be shift-variant. However,  the fusion process does not need to involve the convolution operator $H$ if the objects are in-focus in one of the image pair (e.g. clocks), because the higher values of the sensor gain $\beta$ will be assigned to the pixels of the image that are sharper. That is, a sharp result is obtained from the in-focus regions without applying deconvolution.

\begin{figure}[ht!]
\centering
\includegraphics[width=\columnwidth]{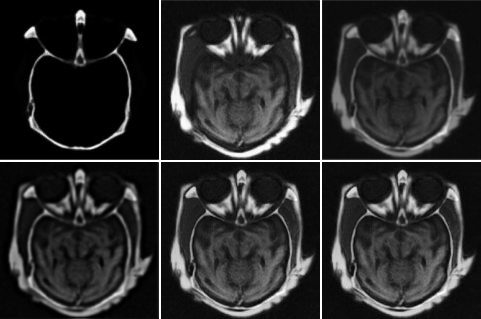}
\caption{Fusion of Head Images. Top-row from left to right: Original CT image, Original MRI, Wavelet-WA Fusion, Bottom-row from left to right: SRDL Fusion, Fusion by $L_1$ norm regularization, GMC regularization}
\label{fig:figure2}
\end{figure}

\begin{figure*}[ht!]
\centering
\includegraphics[width=\textwidth]{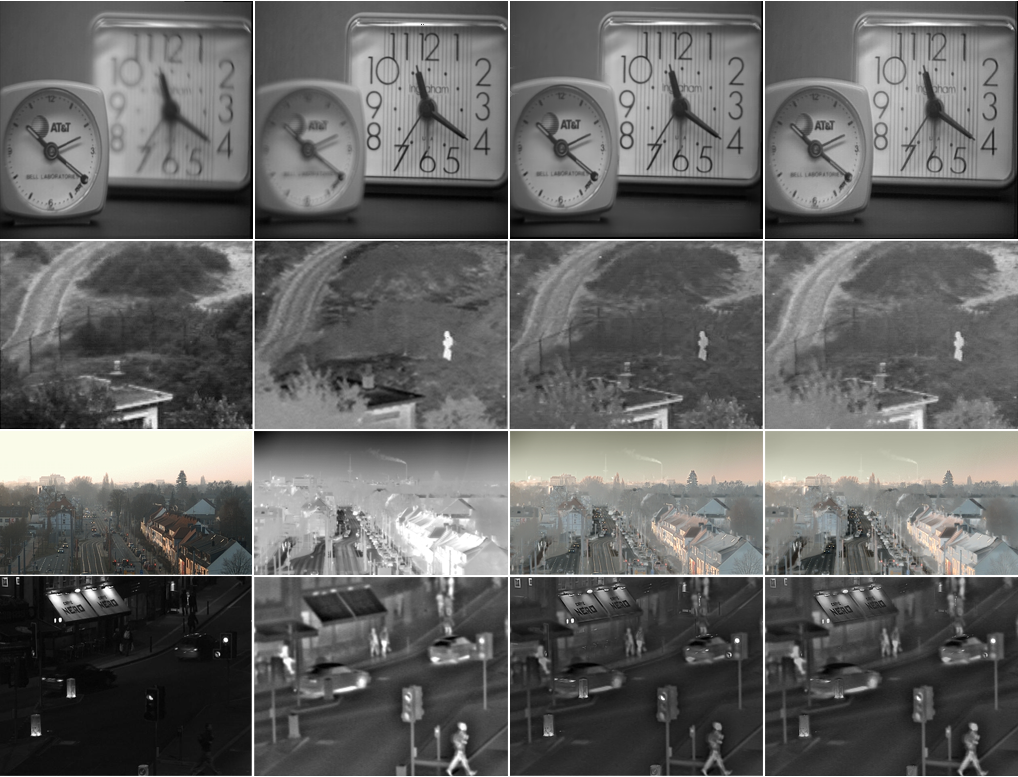}
\caption{Image fusion results of four image pairs: Clock (multi focus), UN camp (visible+IR), Building (colour visible+IR) and Junction (visible+IR). The original image pairs are shown in the two most left columns. The third and the fourth columns are the results of CSR \cite{Liu:fusion:2016} and the proposed GMC method, respectively. }
\label{fig:figure10}
\end{figure*}

Fig. \ref{fig:figure2} shows the fusion results on the human head dataset generated using Wavelet-WA, SRDL, $L_1$ norm and GMC regularization. The proposed method clearly shows better contrast than the others. Fig. \ref{fig:figure10} shows the results of the clock, UN camp, Building and Junction image pairs fused using the top two best methods in terms of objective scores, CSR and GMC methods (see Table \ref{tab:tab1}). The CSR method achieves better contrast in low-intensity areas, but the GMC regularization gives more distinctive heat areas from the IR camera, e.g. where the human is. The results of CSR are also noisier than those of the GMC method.

\subsection{Joint image fusion and deconvolution}
The superiority of GMC regularization can be shown more obviously when the input images are blurred and noisy so that the data fidelity term has a greater effect. Hence, to further assess the merits of our proposed joint image fusion and deconvolution method, in the second experiment, a pair of retinal images, i.e. an optical coherence tomography (OCT) and a fundus image (600$\times$600 pixels), was employed, previously used and described in~\cite{b:ANANTRASIRICHAI2014}. It is obvious that the source images, the OCT in particular, are substantially blurred, hence deconvolution is particularly important in this case. The operator $H$ represents the point spread function (PSF) and for its estimation we use an algorithm proposed in \cite{10} and adapted for B-mode images.
{Generally, different applications require different methods to estimate the PSF. However, this method was proved efficient in \cite{b:achim17a} for ultrasound images, which are similar to OCT. Moreover, in practical scenarios, the PSF for OCT is known.}

\begin{figure*}[ht!]
\centering
\includegraphics[width=\textwidth]{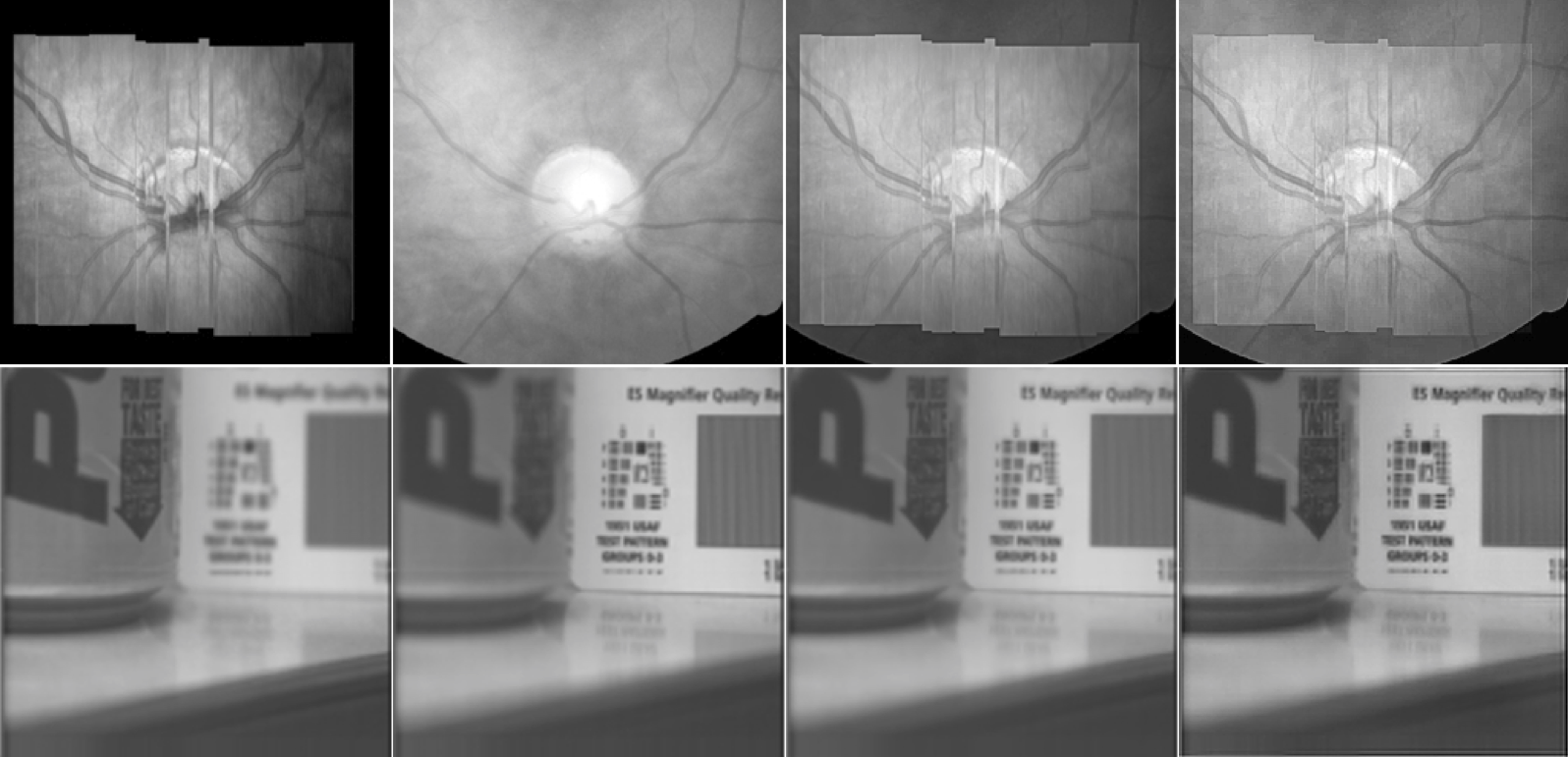}
\caption{Image fusion results of retina (OCT+fundus) image pair (Top row) and enlarged multi-focus image pair, Slika (Bottom row). The original image pairs are shown in the most left columns. The third and the fourth columns are the results of CSR \cite{Liu:fusion:2016} and the proposed GMC method, respectively.}
\label{fig:gather_results_oct}
\end{figure*}

 It can be observed that these images are also affected by noise significantly, hence the ideal value of $\lambda$ was found to be much larger, i.e. $\lambda=0.5$. The results of our joint image fusion and deconvolution algorithms are shown in Fig. \ref{fig:gather_results_oct} top row. In the results, the restored images using the proposed algorithm based on GMC regularization provide the desired visual effects, combining the useful information from the source images while at the same time achieving the desired deblurring effect. {The fused image reveals details from both source images in particular at the optic nerve head (ONH), where the blood vessels and cup size are used to examine some eye diseases, e.g. in glaucoma research  \cite{Bourne:optic:2006}}. This demonstrates the feasibility of the proposed image fusion algorithm with GMC regularization in practical image processing application.
 
We also tested our method with enlarged multi-focus images, Slika, of which the original resolution was 127$\times$127 pixels. The input images were enlarged to the new resolution of 512$\times$512 pixels using Bicubic interpolation technique. Hence, the new image pair does not have sharp areas any more and the deconvolution is required in the fusion process. This is different from the case of the clock images tested in Section \ref{ssec:imagefusion}, where one clock is sharp in one image of the pair thereby not requiring deconvolution.
We simply estimated the PSF using the same method above and set $\lambda=0.005$ because the input images are not noisy. Fig. \ref{fig:gather_results_oct} bottom row shows the fused result of the GMC regularization (4th column). Comparing to the result of the CSR (3th column) and also original image pair (1st and 2nd columns), our fused image is sharper in all areas and has better contrast.

\subsection{Objective assessment}
For objective evaluation, the assessment metrics used here to evaluate fusion performance include Petrovic and Xydeas’s metric ($Pe$) \cite{7}, Wang’s metric ($Q_0$) \cite{8}, and Piella’s metric ($Q$) \cite{9}. Fusion performance assessment remains a difficult task, given that no single assessment metric can provide a full depiction of the merits of one method. Hence, we use a range of measures and aim to draw meaningful conclusions. {Briefly,  Pe measures the perceptual loss and edge preservation value. Q0 measures  loss
of correlation, luminance distortion, and contrast distortion. Q relies on Q0 but with added local salient information.} The scores are shown in Table \ref{tab:tab1}. It can be seen that the proposed image fusion method with $L_1$ norm and GMC sparse regularization have better performance than Wavelet-WA and SRDL algorithms. 
The proposed method compares favourably with CSR, consistently achieving the best $Q$ metric results and sharing the wins in terms of $Pe$. Nevertheless, an important advantage of GMC is that it is 25 times faster than CSR.
When comparing $L_1$ norm and GMC regularization, it can be seen that according to all the assessment metrics, the performance of the latter is superior. Our experiments indicate that the $Pe$ metric is maximised for small values of $\lambda$, high $Q_0$ requires a large $\lambda$ value, while $Q$ metric requires an intermediate $\lambda$. This is somewhat unsurprising since the various measures of image fusion quality have been designed with different objectives in mind.

\begin{table*}
\caption{{Image fusion quality and computational time. The last row groups the average of all images and the average computational time per 10000 pixels.}}
 \centering
 \begin{tabular}{cccccc}
 \hline
  dataset & method & $Pe$ & $Q_0$ & $Q$ & cpu time (sec) \\
	\hline
	\multirow{5}{*}{Clock} & Wavelet-WA \citep{b:achim08b} & 0.5070 & 0.8225 & 0.7122 & 0.10\\
	& SRDL \citep{11} & 0.6864 & 0.9106 & 0.7737 & 408.10\\
	& CSR \citep{Liu:fusion:2016} & \textbf{0.7469} & \textbf{0.9777} & 0.7548 & 151.36 \\
	& {CNN \citep{Ren:infrared:2018}} & 0.7390 & 0.9769 & \textbf{0.8003} & 56.57 \\
	& $L_1$ & 0.6867 & 0.9092 & 0.7231 & 6.20\\
	& proposed GMC & 0.7020 & 0.9125 & 0.7883 & 6.26\\
	\hline
	\multirow{5}{*}{UN camp} & Wavelet-WA \citep{b:achim08b}  & 0.3409 & 0.6235 & 0.6451 & 0.03 \\
	& SRDL \citep{11} & 0.4467 & 0.7004 & 0.7178 & 126.08\\
	& CSR \citep{Liu:fusion:2016}  & 0.4670 & 0.9191 & 0.7203 & 40.08\\
	& {CNN \citep{Ren:infrared:2018}} & \textbf{0.5441} & \textbf{0.9145} & \textbf{0.7280} & 22.79 \\
	& $L_1$ & 0.4409 & 0.6796 & 0.7162 & 1.50\\
	& proposed GMC & 0.4631 & 0.7016 & \textbf{0.7280} & 1.51\\
	\hline
	\multirow{5}{*}{Head} & Wavelet-WA \citep{b:achim08b}  & 0.4050 & 0.6361 & 0.6383 & 0.02 \\
	& SRDL \citep{11} & 0.6890 & 0.7243 & 0.6441 & 33.55\\
	& CSR \citep{Liu:fusion:2016}  & 0.6531 & \textbf{0.9064} & 0.5698 & 28.57\\
	& {CNN \citep{Ren:infrared:2018}} & 0.7865 & 0.8431 & 0.8052 & 11.95 \\
	& $L_1$ & 0.7863 & 0.7943 & 0.8086 & 1.15\\
	& proposed GMC & \textbf{0.7879} & 0.8011 & \textbf{0.8177} & 1.15\\
	\hline
	\multirow{5}{*}{Building} & Wavelet-WA \citep{b:achim08b}  & 0.3252 & 0.6578 & 0.6439 & 0.07\\
	& SRDL \citep{11} & 0.5294 & 0.8988 & \textbf{0.7209 }& 1215.80\\
	& CSR \citep{Liu:fusion:2016}   & \textbf{0.6653} & 0.8053 & 0.5706 & 115.01\\
	& {CNN \citep{Ren:infrared:2018}} & 0.5334 & 0.8916 & 0.6116 & 61.23 \\
	& $L_1$ & 0.4441 & 0.6266 & 0.5661 & 5.75\\
	& proposed GMC & 0.5733 & \textbf{0.8994} & 0.6519 & 6.02\\
	\hline
	\multirow{5}{*}{Junction} & Wavelet-WA \citep{b:achim08b}  &  0.3055 & 0.8433 &     0.6143 & 0.06\\
	& SRDL \citep{11} & 0.5450 & 0.9026 & 0.7084 & 937.6\\
	& CSR \citep{Liu:fusion:2016}   &     0.4886 & 0.9020 &    0.6531 & 96.82\\
	&{CNN \citep{Ren:infrared:2018}} & \textbf{0.5879 }& 0.8947 & \textbf{0.7752} & 54.13 \\
	& $L_1$ &  0.4696  & 0.8670 &     0.7131 &      1.53\\
	& proposed GMC &   {0.5793} & \textbf{0.9114} &    0.7186 & 3.86\\
	
	\hline
	\multirow{5}{*}{Octec} & Wavelet-WA \citep{b:achim08b}  & 0.6480 & 0.9365 &   0.4027 & 0.05\\
	& SRDL \citep{11} & 0.5254 & 0.9516 & \textbf{0.6624 }& 1022.18\\
	& CSR \citep{Liu:fusion:2016}   &   0.6932& \textbf{0.9556}&       0.5373 & 82.03\\
	& {CNN \citep{Ren:infrared:2018}} & 0.5682 & 0.9390 & 0.6274 & 55.67 \\
	& $L_1$ &   0.6996& 0.9336 &    0.4702        &      3.67\\
	& proposed GMC &   \textbf{0.7145 }& 0.9409   &      0.6039 & 4.35\\
	
	\hline
	\multirow{5}{*}{Retina} & Wavelet-WA  \citep{b:achim08b}  & 0.3113 & 0.6472 & 0.6302 & 0.17\\
	& SRDL \citep{11} & 0.6011 & 0.7098 & 0.6321 & 526.85\\
	& CSR \citep{Liu:fusion:2016} & 0.5221 & \textbf{0.9664} & 0.6486 & 190.74 \\
	& {CNN \citep{Ren:infrared:2018}} & 0.3900 & 0.8009 & 0.4882 & 86.95 \\
	& $L_1$ & 0.4722 & 0.6272 & 0.5730 & 7.80 \\
	& proposed GMC & \textbf{0.6459} & 0.7283 & \textbf{0.6495} & 7.82 \\
\hline
\multirow{5}{*}{Slika} & Wavelet-WA  \citep{b:achim08b}  & 0.7129 & 0.9738 & 0.6556  & 0.08\\
	& SRDL \citep{11} & 0.7538 & 0.9763 & 0.6925 & 673.84\\
	& CSR \citep{Liu:fusion:2016} & 0.7433 & \textbf{0.9850} & 0.6519 & 51.80 \\
	& {CNN \citep{Ren:infrared:2018} }& \textbf{0.7573} & 0.9807 & 0.6616 & 55.15 \\
	& $L_1$ & 0.7089& 0.9737 &0.6352 & 3.02 \\
	& proposed GMC & 0.7175 & \textbf{0.9850} & \textbf{0.6975} & 3.69 \\
\hline
\multirow{5}{*}{Average} & Wavelet-WA  \citep{b:achim08b}  & 0.4445 &    0.7676  &  0.6178    & 0.004\\
	& SRDL \citep{11} & 0.5971   &  0.8468 & 0.6940   & 22.49\\
	& CSR \citep{Liu:fusion:2016} & 0.6225   & \textbf{ 0.9272} & 0.6383  & 5.21 \\
	& {CNN \citep{Ren:infrared:2018}} & 0.6133 &   0.9052 &   0.6872 &  2.24 \\
	& $L_1$ & 0.5885&    0.8014 & 0.6507  &   0.20 \\
	& proposed GMC & \textbf{0.6479 }&     0.8600 &\textbf{ 0.7069} &    0.21\\
\hline
 \end{tabular}
 \label{tab:tab1}
\end{table*}

\section{Conclusion}
\label{sec:conclusion}

This paper proposed a novel image fusion algorithm in a variational framework based on sparse representations and sparse regularization using a specific class of non-convex penalty functions (i.e. GMC). The main contributions consist in demonstrating an existing framework for GMC regularization to 2-D images and subsequently addressing the image fusion problem as an inverse one. In addition, the problem of simultaneously fusing and deconvolving images is addressed. The mathematical derivation of the solution is presented and the condition for suitably choosing the penalty function that ensures convexity of the overall cost function is proved. The proposed algorithm shows competitive performance when benchmarked against both classical and modern image fusion algorithms.

\bibliographystyle{model1-num-names}
\bibliography{ren, alin}

\end{document}